\documentclass[letterpaper, 10 pt, conference]{ieeeconf}  

\IEEEoverridecommandlockouts                              

\usepackage{graphics} 
\usepackage{graphicx} 
\usepackage{amssymb}  
\usepackage{subfigure}
\usepackage{caption}
\usepackage{amsmath} 

\usepackage{booktabs} 
\usepackage{multirow} 
\usepackage{adjustbox} 

\usepackage{tabularx}
\usepackage{siunitx}
\newcolumntype{Y}{>{\raggedright\arraybackslash}X}

\usepackage[hidelinks]{hyperref}

\title{ \LARGE \bf VAR-SLAM: Visual Adaptive and Robust SLAM for Dynamic Environments}

\author{ João Carlos Virgolino Soares*, Gabriel Fischer Abati, Claudio Semini
\thanks{All authors are with the Dynamic Legged Systems (DLS) lab, Istituto
Italiano di Tecnologia (IIT), Genova, Italy.}
\thanks{* Corresponding author, {\tt\small joao.virgolino@iit.it}}
}

\begin{document}

\maketitle

\begin{abstract}

Visual SLAM in dynamic environments remains challenging, as several existing methods rely on semantic filtering that only handles known object classes, or use fixed robust kernels that cannot adapt to unknown moving objects, leading to degraded accuracy when they appear in the scene. We present VAR-SLAM (Visual Adaptive and Robust SLAM), an ORB-SLAM3-based system that combines a lightweight semantic keypoint filter to deal with known moving objects, with Barron's adaptive robust loss to handle unknown ones. The shape parameter of the robust kernel is estimated online from residuals, allowing the system to automatically adjust between Gaussian and heavy-tailed behavior. We evaluate VAR-SLAM on the TUM RGB-D, Bonn RGB-D Dynamic, and OpenLORIS datasets, which include both known and unknown moving objects. Results show improved trajectory accuracy and robustness over state-of-the-art baselines, achieving up to 25\% lower ATE RMSE than NGD-SLAM on challenging sequences, while maintaining  performance at 27 FPS on average. The code is available at~\url{https://github.com/iit-DLSLab/VAR-SLAM}.

\end{abstract}

\section{INTRODUCTION}

Simultaneous Localization and Mapping (SLAM) is a fundamental capability for autonomous robots, enabling them to build maps of unknown environments while estimating their own trajectory. Visual SLAM, in particular, has received extensive attention due to the  low cost of cameras. 

Existing visual SLAM methods can be broadly divided into feature-based and dense approaches. Dense methods~\cite{dtam2011} directly optimize pixel intensities, while feature-based methods~\cite{orbslam} rely on keypoint detection and matching. Although dense methods achieve high accuracy in static environments, they are more computationally demanding and brittle under strong appearance changes~\cite{dynaslam}. Feature-based methods are generally more efficient, but most classical approaches were not designed for dynamic environments.

Classical systems such as PTAM~\cite{klein07parallel} and ORB-SLAM~\cite{orbslam} employ robust kernels (e.g., Tukey or Huber~\cite{Zhang1997ParameterEstimation}) during optimization to mitigate outliers. However, these fixed kernels are insufficient to cope with high levels of dynamic motion, where a large fraction of features belongs to moving objects.

 \begin{figure}[t!]
    \centering{\includegraphics[width=1.0\columnwidth]{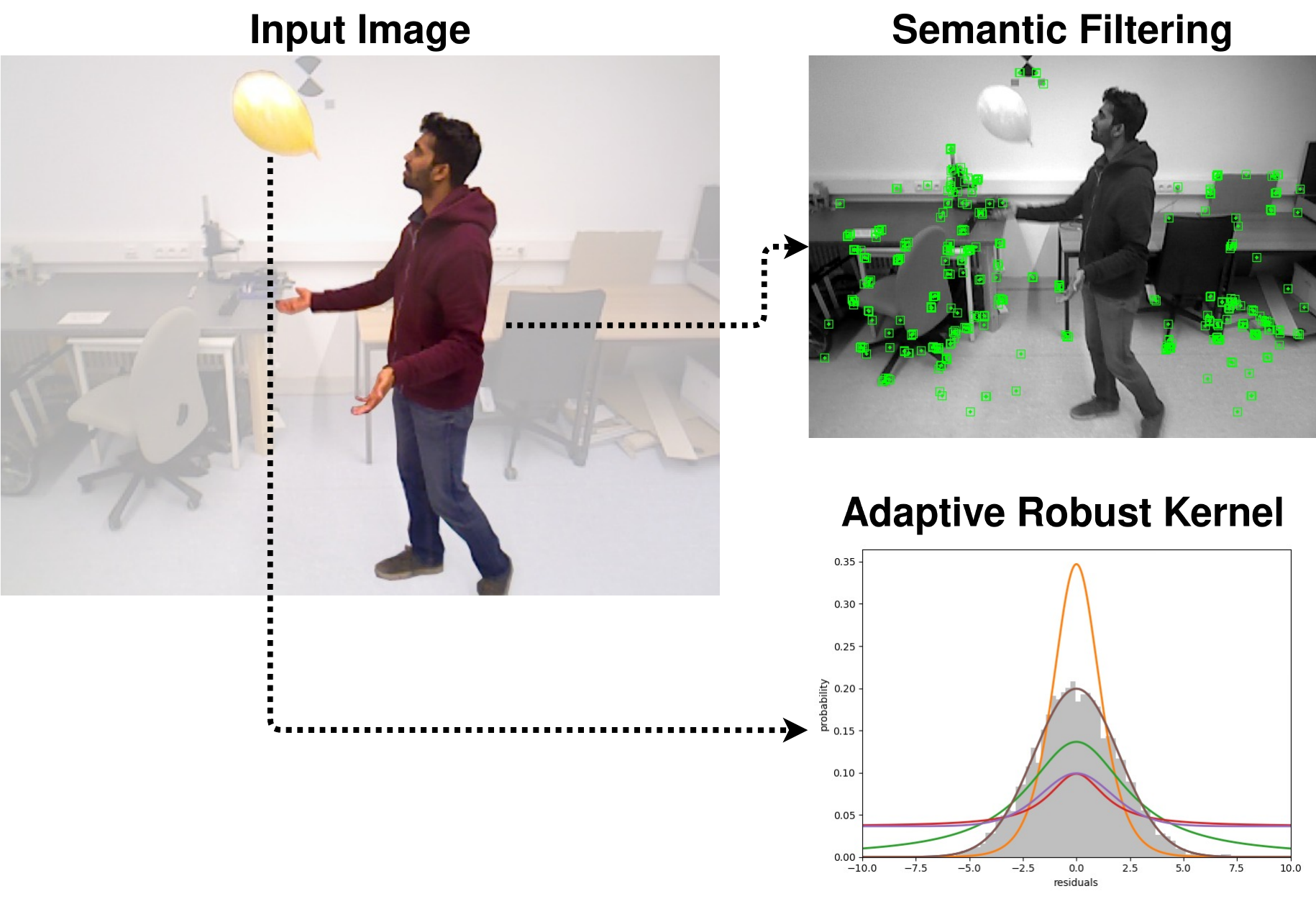}}
          \caption{Overview of our approach. Known dynamic objects (person) are filtered using semantic detection, while unknown dynamic features (balloon) are handled by adaptive robust kernels. Together, these modules enable robust SLAM in dynamic environments.}
  	\label{fig:cover}
\end{figure}

Feature-based SLAM systems have typically addressed dynamics in three ways: (i) geometry-based methods, which detect inconsistent features via epipolar or multi-view constraints; (ii) semantic methods, which leverage deep networks to filter features from known dynamic classes such as people or vehicles~\cite{soares_jint}~\cite{detect-slam}; and (iii) hybrid methods, which combine geometric and semantic cues~\cite{dynaslam}~\cite{dsslam}. Geometry-based methods can, in principle, handle unknown moving objects, but they are often computationally expensive~\cite{sadslam} or involve several hand-tuned thresholds~\cite{ovd-slam2023}. Semantic approaches are efficient but primarily rely on known classes: for example, a carried chair would not be masked if the network was only trained to detect people. Hybrid methods combine the strengths of both, but still rely heavily on manually tuned thresholds, limiting adaptability across scenarios.

Recently, several works have revisited the use of adaptive robust kernels to improve estimation in the presence of outliers~\cite{da-irrk2025}~\cite{Ming2023ARKSLAM}. Unlike fixed kernels, adaptive kernels adjust their influence function according to the observed residual distribution, allowing the optimizer to balance efficiency and robustness automatically. Chebrolu et al.~\cite{Chebrolu2020AdaptiveRK} demonstrated this idea for LiDAR odometry using the Barron family of losses~\cite{Barron2019}, showing improved robustness to dynamics without manual parameter tuning.

In this paper we combine both lines of work: semantics for filtering known dynamic classes, and adaptive robust kernels for handling unknown dynamic objects and outliers, as shown in Fig. \ref{fig:cover}. We propose VAR-SLAM (\textit{Visual Adaptive and Robust SLAM}), a feature-based SLAM system designed for dynamic environments that integrates adaptive Barron kernels into the ORB-SLAM3~\cite{orbslam3} framework. Unlike prior hybrid methods, our approach does not depend on fixed thresholds or specific tuning to deal with unknown moving objects. Instead, the optimizer estimates the kernel shape parameter online from the residuals.

The adaptive kernel formulation is lightweight, allowing our system to maintain performance at an average of 27 FPS, which is competitive with state-of-the-art dynamic SLAM methods.

\subsection{Contributions}

The contributions of the paper can be summarized as follows:

\begin{itemize}

    \item VAR-SLAM: A feature-based visual SLAM system that combines semantic filtering of known dynamic classes with adaptive robust kernels to downweight unknown dynamic objects, integrated into ORB-SLAM3 without reliance on hand-tuned thresholds.
    
    \item To the best of our knowledge, this is the first visual SLAM system to integrate Barron’s adaptive loss into feature-based SLAM for dynamic environments.
    \item Extensive experiments on TUM RGB-D, Bonn RGB-D Dynamic, and OpenLORIS datasets demonstrate that VAR-SLAM consistently improves robustness over ORB-SLAM3 and performs competitively with state-of-the-art dynamic SLAM systems, while achieving high performance. We achieve the best results in 5 sequences of the Bonn RGB-D Dynamic dataset, while improving up to 25\% with respect to NGD-SLAM~\cite{Zhang2024ngdslam}.

\end{itemize}

\section{RELATED WORK}

\subsection{Visual SLAM in Dynamic Environments}

With the rise of deep learning, semantic segmentation has become a common way to filter dynamic regions. Early feature-based approaches that combined geometry methods with deep learning include DynaSLAM~\cite{dynaslam}, DS-SLAM~\cite{dsslam}, and Detect-SLAM~\cite{detect-slam}. DS-SLAM combines a lightweight SegNet~\cite{Badrinarayanan2015SegNetAD} with a moving-consistency (optical-flow/epipolar) test. DynaSLAM, on the other hand, suffers from the demanding computational power of Mask R-CNN~\cite{maskrcnn} to detect objects in the scene, and Detect-SLAM can only deal with classes trained by the neural network.

Several works extend semantic filtering with additional cues. CFP-SLAM~\cite{cfp-slam} integrates YOLOv5 detection with object tracking and a coarse-to-fine static probability model. While effective, it only updates probabilities for features inside bounding boxes and relies on multiple hand-tuned parameters and thresholds. It reports $\sim$23 FPS in the full configuration.

Panoptic-SLAM~\cite{Panoptic-SLAM} and Zhu et al.~\cite{zhu2022fusingpanoptic} reason about unknown moving objects via panoptic segmentation and epipolar checks, but they require running a panoptic net every frame, which is typically heavier than detection. RDS-SLAM~\cite{rds-slam} achieves higher speed by decoupling the semantic thread, but remains limited to the classes handled by the network. Ji et al.~\cite{ji2021} reduce computational cost by segmenting only keyframes and use clustering and reprojection to handle unknown objects, trading some accuracy for speed.

 OVD-SLAM~\cite{ovd-slam2023}  avoids predefined dynamic labels, using optical-flow consistency to decide which segmented boxes are dynamic, but still depends on object detection, i.e., points outside boxes are treated as static. DFS-SLAM~\cite{dfs-slam} removes a-priori dynamic classes via Mask R-CNN and detects dynamic points with a multi-view depth-difference test, but it is reported to run nearly three times slower than ORB-SLAM2~\cite{orbslam}, even with GPU acceleration.

RSO-SLAM~\cite{qin2024rso-slam} explicitly targets objects not captured by the segmentation network, achieving 22 FPS on GPU and good accuracy in scenes with unknown dynamics. However, it still relies on thresholding and parameter tuning.

NGD-SLAM~\cite{Zhang2024ngdslam} achieves real-time dynamic SLAM on CPU by combining YOLO-based detection with mask propagation, but its robustness is restricted to known dynamic classes and it does not have a method to handle unseen moving objects.

In summary, semantic-based methods are strong when dynamics are covered by the detector, but either become computationally expensive~\cite{dynaslam},~\cite{Panoptic-SLAM}~\cite{zhu2022fusingpanoptic}~\cite{dfs-slam} or do not explicitly handle unknown moving objects~\cite{detect-slam}~\cite{ovd-slam2023}~\cite{Zhang2024ngdslam}~\cite{cfp-slam}~\cite{rds-slam}. Some methods that combine semantics with additional geometric cues~\cite{dsslam}~\cite{qin2024rso-slam} reduce this gap, but often rely on carefully tuned thresholds and cannot adapt their robustness online.

\subsection{Visual SLAM using Robust Kernels}

Robust loss functions have long been used in SLAM optimization to mitigate the effect of outliers. Classical systems such as ORB-SLAM3~\cite{orbslam3} adopt a fixed Huber kernel in bundle adjustment (BA), which improves stability in the presence of mismatches but cannot cope with high levels of dynamic motion where large portions of the scene are in motion.

Recent works have explored adaptive or learned kernels. DynaVINS~\cite{dynavins} combines motion constraints with robust kernels in a visual–inertial framework, improving resilience against dynamic features. 

AEROS~\cite{aeros2022} uses Barron to adapt the robust loss for graph-based SLAM, targeting loop-closure outliers. ARK-SLAM~\cite{Ming2023ARKSLAM} proposes adaptive-kernel BA inside a full monocular visual SLAM method, improving robustness over fixed Huber and Cauchy baselines, but it is not designed for dynamic scenes.

DGO-VINS~\cite{DGO-VINS} uses an adaptive loss function guided by per-feature motion probability, effectively downweighting dynamic features in visual-inertial SLAM, but requiring explicit motion probability estimation rather than per-residual adaptivity. It adapts the residual weighting, not the form of the loss function.

Chebrolu et al.~\cite{Chebrolu2020AdaptiveRK} adapt Barron’s generalized robust loss~\cite{Barron2019} for non-linear least squares, applying it to LiDAR odometry and estimating the shape parameter online from residual distributions. Their results showed improved robustness to outliers without manual tuning.

To the best of our knowledge, no prior feature-based visual SLAM system for dynamic environments integrates Barron’s adaptive loss directly into BA and frame-to-frame tracking and mapping. Most prior art either keeps fixed kernels (Huber/Cauchy), adapts weights rather than the loss shape, or applies adaptivity at the pose-graph level. VAR-SLAM fills this gap by coupling a lightweight semantic keypoint filter with online shape estimation in ORB-SLAM3’s BA.

\section{Mathematical Formulation}

Let $\theta$ denote the set of parameters to be estimated (camera poses and 3D map points), and let $\mathbf{r}_i(\theta)$ be the 2D reprojection residual associated with observation $i$. We define the scalar residual magnitude $e_i(\theta)=\|\mathbf{r}_i(\theta)\|_2$. Robust estimation introduces a loss function $\rho(\cdot)$ to reduce the influence of outliers:

\begin{equation}
\theta^* = \arg \min_\theta \sum_{i=1}^N \rho(e_i(\theta)).
\end{equation}

The choice of $\rho$ determines how strongly large residuals are downweighted. Quadratic losses ($\rho(e)=\tfrac{1}{2}e^2$) treat all errors equally (efficient when noise is Gaussian), whereas robust losses reduce the influence of large residuals to preserve estimator stability under outliers.

\subsection{Barron’s Generalized Robust Kernel}

To handle outliers in optimization, we employ the generalized robust loss introduced by Barron~\cite{Barron2019}, which unifies many classical kernels under a single parametric form. Given a residual $e$, the loss is defined as:

\begin{equation}
\resizebox{\columnwidth}{!}{$
\rho(e;\alpha,c) = 
\begin{cases}
\dfrac{|\alpha - 2|}{\alpha} \left[ \left( \dfrac{e^2}{|\alpha - 2|c^2} + 1 \right)^{\alpha/2} - 1 \right], & \alpha \neq 0,2 \\[0.6em]
\dfrac{e^2}{2c^2}, & \alpha = 2\ (\text{L2}) \\[0.6em]
\log\!\left( \dfrac{e^2}{2c^2} + 1 \right), & \alpha = 0\ (\text{Cauchy}) \\[0.6em]
1 - \exp\!\left(-\dfrac{e^2}{2c^2}\right), & \alpha \to -\infty\ (\text{Welsch})
\end{cases}
$}
\label{eq:barronloss}
\end{equation}

\noindent where, $\alpha \in (-\infty, 2]$, controls the shape of the kernel, and $c$ is a scale parameter. The loss smoothly interpolates between classical estimators. Smaller $\alpha$ increases robustness to heavy-tailed residuals.

\subsection{Negative Log-Likelihood Formulation}

Following Chebrolu et al.~\cite{Chebrolu2020AdaptiveRK}, we interpret Barron’s loss in a probabilistic framework. 
The negative log-likelihood of residuals $\{ e_i(\theta) \}_{i=1}^N$ is:

\begin{equation}
L(\alpha) = \sum_{i=1}^N \left[ \log \tilde{Z}(\alpha) + \rho\big(e_i(\theta); \alpha, c\big) \right],
\end{equation}

where $\tilde{Z}(\alpha)$ is the partition function ensuring normalization. 
The optimal shape parameter is then:

\begin{equation}
\label{eq:nll}
\alpha^* = \arg \min_\alpha L(\alpha).
\end{equation}

Since $\tilde{Z}(\alpha)$ has no closed form, we approximate the solution using a grid search 
over $\alpha \in [-10, 2]$ with step size $0.1$. 
Values of $\log \tilde{Z}(\alpha)$ are precomputed and stored in a lookup table for efficiency.

\section{METHODOLOGY}

\begin{figure}[t!]
    \centering{\includegraphics[width=1.0\columnwidth]{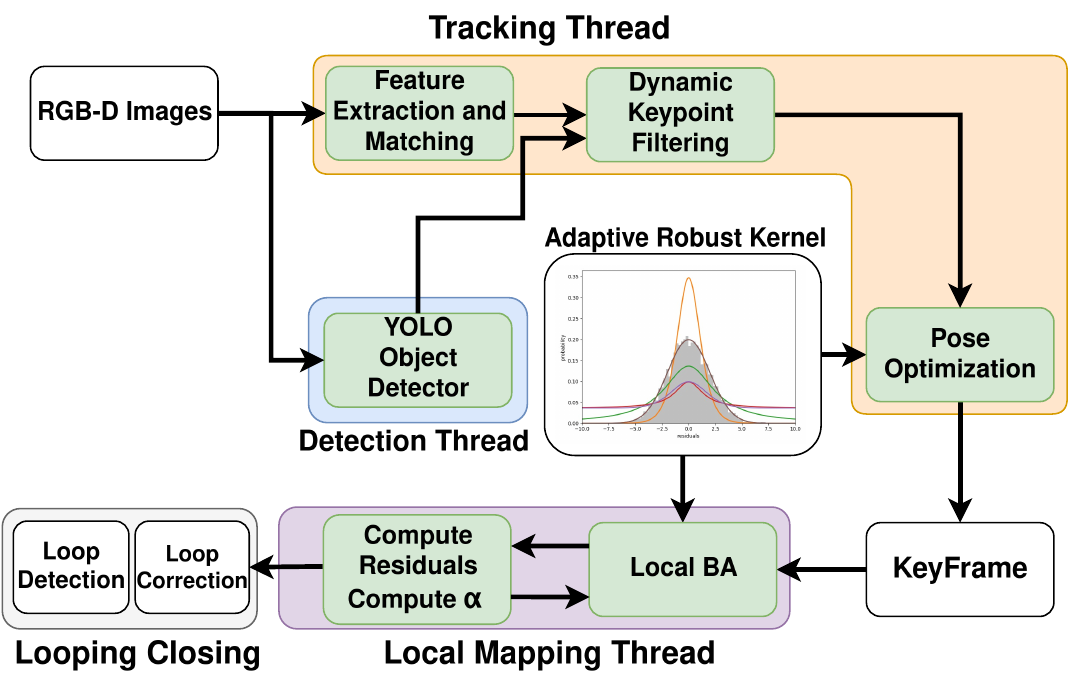}}
          \caption{System framework. Our method extends ORB-SLAM3 by introducing semantic filtering to remove features on known dynamic objects, and adaptive robust kernels for handling unknown dynamics. Pose optimization uses a fixed $\alpha$=1 for stable tracking and efficiency, while Local BA employs alternating optimization with the Barron kernel. Global BA is left unchanged.}
  	\label{fig:methodology}
\end{figure}

VAR-SLAM combines lightweight semantic filtering of \emph{known} dynamic classes with adaptive robust optimization to handle \emph{unknown} moving objects and other outliers. Fig.~\ref{fig:methodology} illustrates the proposed method. First, an object detector removes keypoints on known dynamic classes (people in our implementation). Second, the remaining measurements are optimized using Barron’s loss. In Local BA we estimate the shape parameter $\alpha$ online from the current residuals, whereas frame-to-frame pose optimization uses a fixed $\alpha$ to ensure stability and performance.

\subsection{Dynamic Feature Filtering}

\begin{figure}[h!]

  \subfigure[]{%
    \includegraphics[width=.3\linewidth]{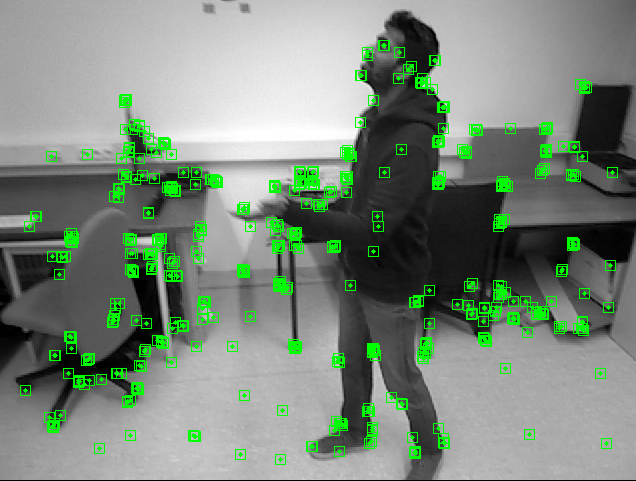}
    \label{fig:keypoin1}
  } 
  \subfigure[]{%
    \includegraphics[width=.3\linewidth]{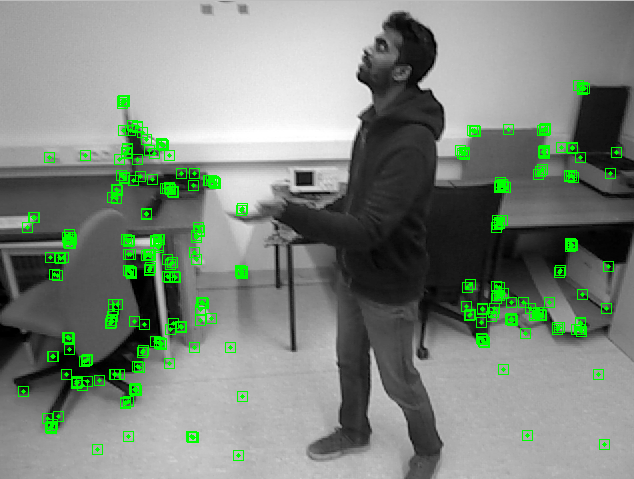}
    \label{fig:keypoin2} 
  }
  \subfigure[]{%
    \includegraphics[width=.3\linewidth]{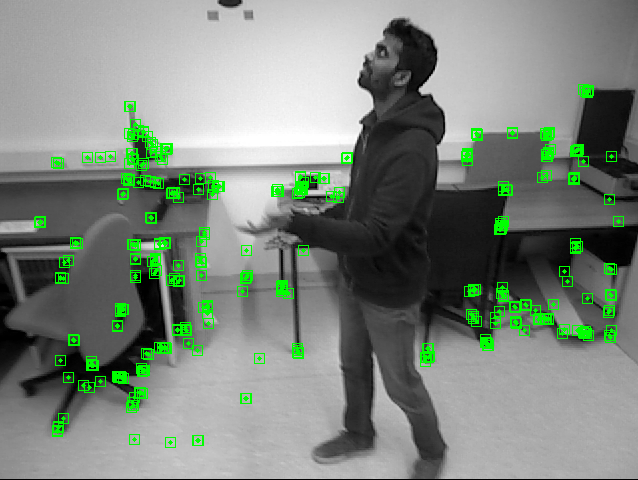}
    \label{fig:keypoint3}
  } 
  \caption{Dynamic keypoint filtering. (a) ORB keypoints before filtering. (b) keypoints filtered inside the bounding box from YOLOv4 detection. (c) keypoint removal with depth-awareness: keypoints whose depth is consistent with the foreground person are discarded.} 
  \label{fig:keypointfiltering}
\end{figure}

Dynamic objects (e.g., people) are a major source of outliers. While robust kernels downweight large residuals, optimization can still be biased if a dominant fraction of features lies on moving objects. A pre-trained YOLOv4~\cite{yolov4} detector is used to obtain bounding boxes for known dynamic classes (people, in our implementation). To avoid removing useful background structure that falls inside these boxes, we adopt the strategy of~\cite{Changing-SLAM}: for keypoints inside a detection box, we use their RGB-D depth to keep those that do not belong to the person (i.e., whose depth is inconsistent with the foreground), and discard the rest. This yields a segmentation-like effect using only detections and depth. The three stages of dynamic keypoint filtering are shown in Fig.~\ref{fig:keypointfiltering}.

\subsection{Problem Formulation in ORB-SLAM}

In ORB-SLAM3, camera poses and 3D map points are estimated by minimizing robustified reprojection errors. 
For an observation of a 3D point $\mathbf{X}_i$ in frame $k$ with pose 
$\mathbf{T}_{cw}^k \in \mathrm{SE}(3)$, 

\begin{equation}
\mathbf{r}_{ik}(\theta) \;=\; \mathbf{u}_{ik} - \pi\!\big(\mathbf{T}_{cw}^k \mathbf{X}_i\big), 
\qquad
e_{ik} \;=\; \|\mathbf{r}_{ik}(\theta)\|_2.
\end{equation}

\noindent where $\mathbf{u}_{ik}$ is the measured keypoint location and $\pi(\cdot)$ is the projection function that maps 3D camera coordinates into 2D image coordinates.

The original system minimizes a sum of Huber losses (applied to squared residuals) across all observations:

\begin{equation}
\theta^* \;=\; \arg\min_{\theta}\; \sum_{i,k} \rho_{\text{Huber}}\!\big(e_{ik}^2\big).
\end{equation}

While effective in moderately noisy conditions, the Huber kernel cannot adapt to residual distributions dominated by dynamic objects.

\subsection{Adaptive Estimation of $\alpha$}

We estimate $\alpha$ by alternating minimization over $\{\theta,\alpha\}$:

\begin{enumerate}
\item \textbf{Fix $\theta$, update $\alpha$:} 
\(
\alpha^{(t)} \leftarrow \arg\min_{\alpha} L(\alpha)
\)
using~\eqref{eq:nll} on the current active residuals.
\item \textbf{Fix $\alpha$, update $\theta$:}
\(
\theta^{(t)} \leftarrow \arg\min_{\theta} \sum_i \rho\!\big(e_i(\theta);\alpha^{(t)},c\big),
\)
solved with IRLS in \texttt{g2o} \cite{g2o}.
\end{enumerate}

In static datasets $\alpha$ typically remains near 2, whereas in dynamic datasets it decreases toward 1 or below, reflecting stronger downweighting of outliers.

\subsection{Pose Optimization - Tracking Thread}

In ORB-SLAM3, the camera pose of the current frame $k$ is refined by minimizing reprojection errors 
while keeping associated map points fixed. In VAR-SLAM, we replace the Huber kernel with Barron’s generalized robust loss: 

\begin{equation}
\mathbf{T}_{cw}^{k*} 
= \arg\min_{\mathbf{T}_{cw}^k} 
\sum_{i \in \mathcal{P}_k}
\rho\!\left( \big\| \mathbf{u}_{ik} - \pi(\mathbf{T}_{cw}^k \mathbf{X}_i) \big\|^2 \,;\, \alpha, c \right).
\end{equation}

\noindent where $\mathcal{P}_k$ denotes the set of map points observed in frame $k$, 
$\mathbf{u}_{ik}$ is the keypoint measurement, and $\pi(\cdot)$ is the projection function. To ensure stable tracking, we fix $\alpha = 1$.

\subsection{Local Bundle Adjustment - Local Mapping Thread}

For each new keyframe, LocalBA jointly optimizes the poses of the local window and associated map points. Here, residuals vary significantly due to dynamics and occlusions. We therefore apply the alternating minimization scheme: $\alpha$ is updated at each iteration based on residuals, while poses and points are refined in g2o. This allows robustness to adapt dynamically to the local residual distribution.

Local BA jointly optimizes the poses of a sliding window of keyframes $\mathcal{K}$ 
and the map points they observe. In VAR-SLAM, the robust kernel is replaced with Barron’s loss, and the shape parameter 
$\alpha$ is optimized jointly with the state:

\begin{equation}
\begin{aligned}
\{\mathbf{T}_{cw}^j, \mathbf{X}_i, \alpha\}^* 
&= \arg\min_{\{\mathbf{T}_{cw}^j\}, \{\mathbf{X}_i\}, \alpha}
\sum_{j \in \mathcal{K}} \sum_{i \in \mathcal{P}_j} \\
&\quad \rho\!\left(
\| \mathbf{u}_{ij} - \pi(\mathbf{T}_{cw}^j \mathbf{X}_i) \|^2 ; \alpha, c
\right).
\end{aligned}
\end{equation}

\noindent where $\mathcal{P}_j$ denotes the set of map points observed in keyframe $j$, 
$\mathbf{u}_{ij}$ is the measured keypoint, and $\pi(\cdot)$ is the projection function. We fix the scale parameter to $c=1$  to decouple shape and scale, and avoid $\alpha–c$ ambiguity~\cite{Chebrolu2020AdaptiveRK}.

\section{RESULTS}

This section presents quantitative results on real-world datasets widely used in the SLAM community. We evaluate on TUM RGB-D~\cite{benchmark}, using the highly dynamic sequences from the fr3/walking subset (xyz, rpy, halfsphere and static). We also include the Bonn RGB-D Dynamic dataset~\cite{re-fusion}, which contain scenes with objects unknown a priori to the detector (e.g., cardboard box, balloon). We additionally report results on two OpenLORIS~\cite{openloris} sequences (Market and Office) to assess domain shift. We use the root mean square error (RMSE) of the Absolute Trajectory Error (ATE)~\cite{benchmark} as the evaluation metric to compare the methods.

\subsection{TUM RGB-D Dataset}

For the TUM RGB-D dataset~\cite{benchmark}, we compared our method against several state-of-the-art feature-based visual SLAM methods from the literature, including DynaSLAM~\cite{dynaslam}, DS-SLAM~\cite{dsslam}, RDS-SLAM~\cite{rds-slam}, CFP-SLAM~\cite{cfp-slam}, and NGD-SLAM~\cite{Zhang2024ngdslam}. 

Table~\ref{tab:TUM} summarizes the results on the four fr3/walking sequences.
On \texttt{fr3\_w\_xyz}, VAR-SLAM obtained 2.2 cm, close to the best methods (1.5 cm for DynaSLAM, CFP-SLAM and NGD-SLAM) while running at $\sim$27 FPS. 

On \texttt{fr3\_w\_rpy} and \texttt{fr3\_w\_half}, VAR-SLAM achieves 5.9 cm and 4.6 cm, respectively, within 1–2.5 cm of the best real-time methods (NGD-SLAM and CFP-SLAM). On \texttt{fr3\_w\_static}, all approaches are essentially tied (0.7–0.8 cm).

The TUM walking sequences contain only people as dynamic objects. Therefore, methods that rely heavily on person segmentation, dense masks, or mask propagation (e.g., DynaSLAM, CFP-SLAM  and NGD-SLAM) obtain lower RMSE. In contrast, VAR-SLAM maintain near real-time operation and focuses on robustness to unknown moving objects via the adaptive kernel. This design choice pays off on the Bonn RGB-D Dynamic dataset, with more diverse scenarios, while remaining competitive on TUM.

\begin{table*}[h]
\centering
\caption{Comparison of RMSE of the ATE (cm) on the TUM RGB-D dataset. The best results are highlighted in bold and the second-best are underlined. The symbol * indicates methods that are real-time or near real-time.}
\label{tab:ate_comparison}
\begin{tabular}{lcccccc}
\hline
Sequence & DynaSLAM & DS-SLAM & RDS-SLAM &  CFP-SLAM* & NGD-SLAM* & VAR-SLAM* (Ours) \\ 
\hline
f3/w\_xyz  & \textbf{1.5} & 2.4 & 5.7   & \textbf{1.5} & \textbf{1.5} & \underline{2.2} \\
f3/w\_rpy  & \underline{3.6} & 44.4 & 16.0   & 4.1 & \textbf{3.4} & 5.9 \\
f3/w\_half  & 2.7 & 3.0 & 8.1   & \textbf{2.3} & \underline{2.4}  & 4.6 \\
f3/w\_static & \textbf{0.7} & \underline{0.8} & \underline{0.8}   & \textbf{0.7} & \textbf{0.7} & \underline{0.8} \\
\hline
\end{tabular}%
\label{tab:TUM}
\end{table*}

\subsection{Bonn RGB-D Dynamic Dataset}

The Bonn dataset comprises three families of sequences. In the first group, comprising \texttt{crowd} and \texttt{people\_tracking} sequences, there are mainly people moving in front of the camera. The second group contain people moving and interacting with small or medium objects that do not occupy a large portion of the image, e.g., balloons and cardboard boxes, being moved in the scene. These objects are usually not present in the pre-trained models of object detectors and segmentation networks~\cite{coco}. This group includes the \texttt{balloon} and variances of the \texttt{nonobs} sequences.

\begin{figure}[h!]
\centering
  \subfigure[]{%
    \includegraphics[width=.47\linewidth]{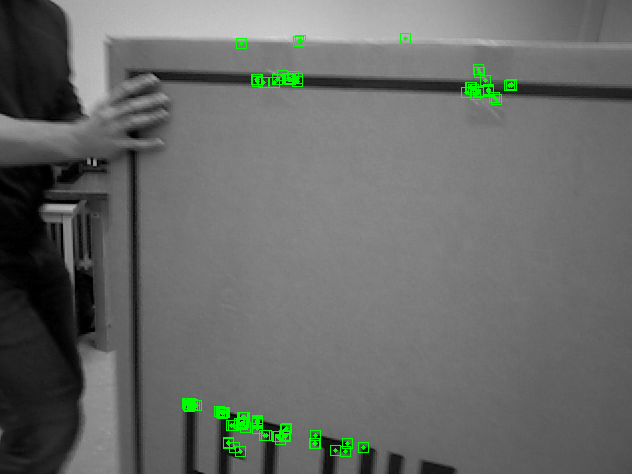}
    \label{fig:obsbox1}
  } 
  \subfigure[]{%
    \includegraphics[width=.47\linewidth]{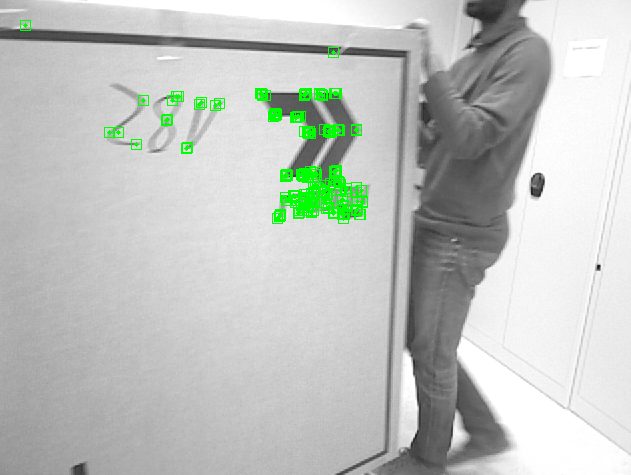}
    \label{fig:obsbox2} 
  }
  \caption{Feature extraction examples from the moving\_obs\_box sequence, where a large box occupies a large portion of the image.} 
  \label{fig:obsbox}
\end{figure}

The third group consists of large moving objects close to the camera that occupy a large portion of the scene, including the sequences \texttt{moving\_obs\_box} and \texttt{placing\_obs\_box}. Fig.~\ref{fig:obsbox} shows examples of feature extraction in scenes from the \texttt{moving\_obs\_box} sequence, where a large portion of the image is occupied by a moving box, causing most of the features to be outliers.

VAR-SLAM is benchmarked against seven feature-based visual SLAM methods from the literature, namely ORB-SLAM3~\cite{orbslam3}, DynaSLAM~\cite{dynaslam}, OVD-SLAM~\cite{ovd-slam2023}, Panoptic-SLAM~\cite{Panoptic-SLAM}, RDS-SLAM~\cite{rds-slam}, RSO-SLAM~\cite{qin2024rso-slam}, and NGD-SLAM~\cite{Zhang2024ngdslam}. 
Tables~\ref{tab:ate_comparison_bonn_people} and~\ref{tab:bonn_ate_comparison} show the comparison of the RMSE ATE between the methods. VAR-SLAM is frequently first or second on sequences with unknown moving objects (e.g., \texttt{moving\_nonobs\_box2}) and remains competitive elsewhere, while RSO-SLAM lead in scenes where the movement of people is dominant (\texttt{crowd}, \texttt{balloon} and \texttt{person\_track}). VAR-SLAM achieves the best ATE in 5 sequences, and second best in 2, out of the 11 evaluated sequences with unknown moving objects. 

\begin{figure}[h!]
\centering
  \subfigure[]{%
    \includegraphics[width=.475\linewidth]{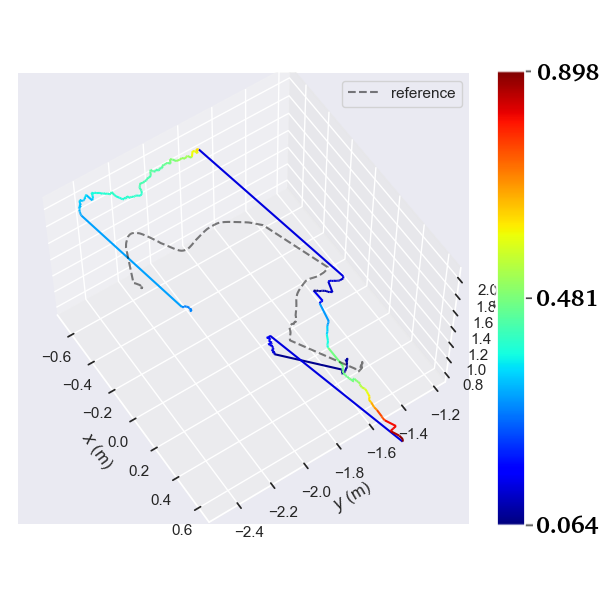}
    \label{fig:traj_moving_obs_NGD}
  } 
  \subfigure[]{%
    \includegraphics[width=.475\linewidth]{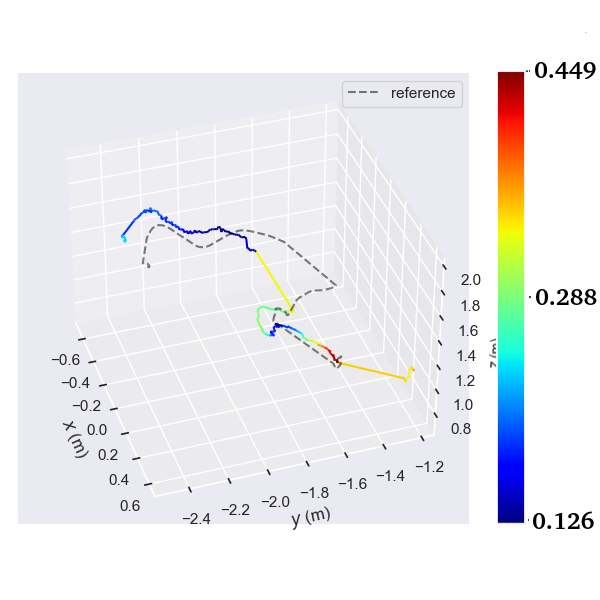}
    
    \label{fig:traj_moving_obs_VAR} 
  }
  \caption{Output trajectories of (a) NGD-SLAM and (b) VAR-SLAM in the \texttt{moving\_obs\_box} sequence.} 
  \label{fig:traj_moving_obs}
\end{figure}

Fig.~\ref{fig:traj_moving_obs} shows that VAR-SLAM halves the peak ATE in the \texttt{moving\_obs\_box}, with respect to the trajectory estimated by NGD-SLAM. Also, it yields a trajectory closer to the reference during heavy occlusion, showing the ability of VAR-SLAM to adapt to scenarios with a high amount of outliers.

\begin{table*}[h]
\centering
\caption{Comparison of RMSE of the ATE (cm). The best results are highlighted in bold and the second-best are underlined. The symbol * indicates methods that are real-time or near real-time.}
\label{tab:ate_comparison_bonn_people}
\begin{tabular}{lcccccc}
\hline
Sequence & ORB-SLAM3* & DynaSLAM & OVD-SLAM &  RSO-SLAM & NGD-SLAM* & VAR-SLAM* (Ours) \\ 
\hline
crowd1 & 45.75 & \textbf{1.6} & 1.8   & \underline{1.66} &  2.4 & \underline{1.66} \\
crowd2 & 91.91 & 3.1 & \underline{2.3}   & \textbf{2.29} & 2.50 & 2.98 \\
crowd3 & 62.27 & 3.8 & \textbf{2.4}   & 3.10 & 3.30  & \underline{2.74} \\
person\_track & 74.17 & 6.1 & \underline{4.10}   & \textbf{3.66} & 4.6 & 4.69 \\
person\_track2 & 80.08 & 7.8 & \underline{3.80}   & \textbf{3.70} & 6.2 & 7.92 \\
moving\_nonobs\_box & 50.88 & 23.2 & \textbf{1.8}   & 2.01 & \underline{1.97} & 2.5 \\
moving\_nonobs\_box2 & 3.72 & 3.9 & \underline{3.3}   & --- & 4.23 & \textbf{2.88} \\
\hline
\end{tabular}%
\end{table*}

\begin{table}[t]
\centering
\caption{Comparison of RMSE of the ATE (cm) on the Bonn RGB-D Dynamic dataset. The best results are highlighted in bold and the second-best are underlined. The symbol * indicate methods that are real-time or near real-time.}
\label{tab:bonn_ate_comparison}
\setlength{\tabcolsep}{1.5pt}
\renewcommand{\arraystretch}{1.2}
\resizebox{\linewidth}{!}{%
\begin{tabular}{|c|c|c|c|c|c|}
\hline
Sequence & ORB-SLAM3* & Panoptic-SLAM & RSO-SLAM & NGD-SLAM* & Ours* \\ 
\hline
placing\_nonobs\_box           & 80.67 & 4.40 & \textbf{1.55} & \underline{1.91} & 2.62 \\
placing\_nonobs\_box2          & 2.31 & 2.20 & ---  & \underline{1.84} & \textbf{1.74} \\
placing\_nonobs\_box3          & 16.10 & --- & \underline{2.5}  & 2.65 & \textbf{2.26} \\
removing\_nonobs\_box            & 1.73 & --- & \textbf{1.53} & 1.57 & \underline{1.56} \\
removing\_nonobs\_box2           & \underline{2.05} & --- & 2.06 & 3.6   & \textbf{2.03} \\
balloon                & 7.12 & 2.90 & \textbf{2.48} & \underline{2.63} & 2.87 \\
balloon2               & 22.38 & 2.70 & \textbf{2.44} & \underline{2.54} & 2.55 \\
moving\_nonobs\_box        & 50.88 & 2.70 & \underline{2.01} & \textbf{1.97} & 2.50 \\
moving\_nonobs\_box2       & 3.72 & \underline{3.30} & ---  & 4.23 & \textbf{2.88} \\
moving\_obs\_box       & 47.82 & --- & \underline{26.55} & 33.99 & \textbf{25.55} \\
placing\_obs\_box           & 33.83 & --- & 24.82 & \textbf{23.22}     & \underline{23.43} \\
\hline
\end{tabular}%
}
\end{table}

\subsection{OpenLORIS Dataset}

We evaluate on two representative sequences from the OpenLORIS dataset~\cite{openloris}, one from the market and one from the office sequences, to test generalization under domain shift (illumination, crowd level) and different layouts. The office sequences are essentially static, while the market sequences are highly cluttered and dynamic, with unknown moving objects (e.g., market carts), and moving people.

To isolate the effect of the robust kernel, we compare ORB-SLAM3 with our method without semantic filtering, denoted VAR-SLAM-RK. Table~\ref{tab:openloris_full} shows the ATE RMSE results, together with the correctness rate (CR), i.e., the percentage of the trajectory completed without losing track.  On the Market scenes, VAR-SLAM-RK reduces the ATE by 15.5\% (\texttt{Market-1}) and 7.6\% (\texttt{Market-3}) relative to ORB-SLAM3, maintaining 100\% CR on Market-1 and a 91.3\% CR on the harder Market-3.

On the \texttt{Office-1} sequence, ORB-SLAM3 already achieves very low error (7.5 cm). Our method yields a small absolute increase (to 9.1 cm) with CR unchanged. Overall, the proposed method improves the accuracy in dynamic, cluttered scenes.

\begin{table}[ht]
\centering
\caption{Evaluation of methods on the OpenLORIS dataset. 
ATE RMSE in meters. Correct Rate in \%.}
\label{tab:openloris_full}

\begin{adjustbox}{max width=\textwidth}
\begin{tabular}{l *{2}{cc}}
\toprule
\multirow{2}{*}{Sequence} 
  & \multicolumn{2}{c}{ORB-SLAM3} 
  & \multicolumn{2}{c}{VAR-SLAM-RK} \\
\cmidrule(lr){2-3} \cmidrule(lr){4-5}
  & ATE RMSE & CR & ATE RMSE & CR \\
\midrule

\multicolumn{5}{l}{\textbf{Office}} \\
Office-1   & 0.075 & 100 & 0.091 & 100 \\
\midrule
\multicolumn{5}{l}{\textbf{Market}} \\
Market-1   & 3.198 & 100 & 2.702 & 100 \\
Market-3   & 3.644 & 95.9 & 3.367 & 91.3 \\

\bottomrule
\end{tabular}
\end{adjustbox}
\end{table}

\subsection{Adaptive $\alpha$ behavior}

To demonstrate that our method truly adapts online, we log the evolution of the shape parameter $\alpha$ during optimization and analyze its behavior across different datasets. 

\begin{figure}[h!]
\centering    \includegraphics[width=.97\linewidth]{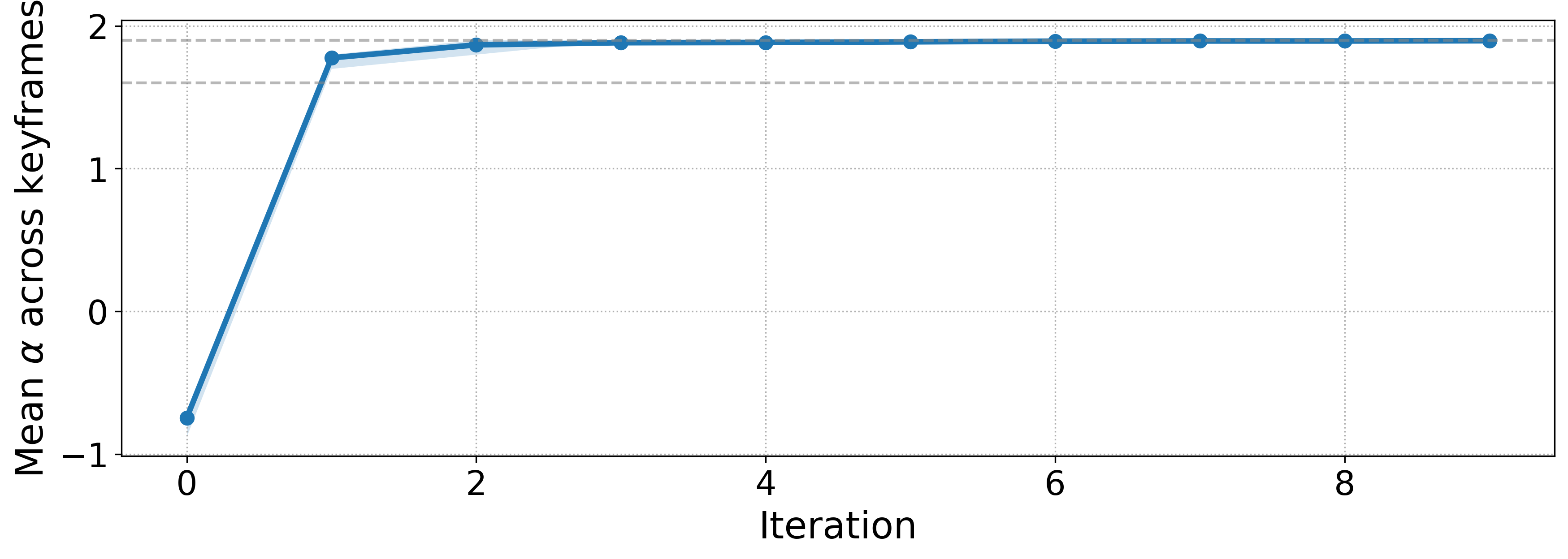}
  \caption{Mean $\alpha$ values across keyframes per iteration in the \texttt{office-1} sequence from the OpenLORIS dataset.} 
  \label{fig:alpha_office}
\end{figure}

The first analysis is on the \texttt{Office-1} sequence from the OpenLORIS dataset. It is a fully static scene, which explains the low ATE obtained by ORB-SLAM3, as shown in Tab.~\ref{tab:openloris_full}. Fig.~\ref{fig:alpha_office} shows the mean $\alpha$ values across keyframes per iteration in the \texttt{Office-1} sequence. It quickly converges to a value close to 2. In a static scene, the residual distribution becomes essentially Gaussian. In this case, the optimizer tends to a standard least squares.  This demonstrates that the adaptive kernel is consistent: it reduces to ORB-SLAM3 behavior when robustness is unnecessary, but automatically switches to heavy-tailed losses in more challenging dynamic environments, as shown in the next example.

\begin{figure}[h!]
\centering
  \subfigure[]{%
    \includegraphics[width=.47\linewidth]{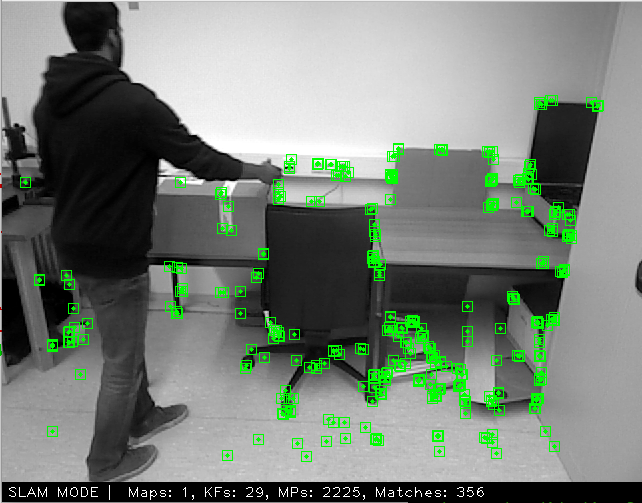}
    \label{fig:KF29}
  } 
  \subfigure[]{%
    \includegraphics[width=.47\linewidth]{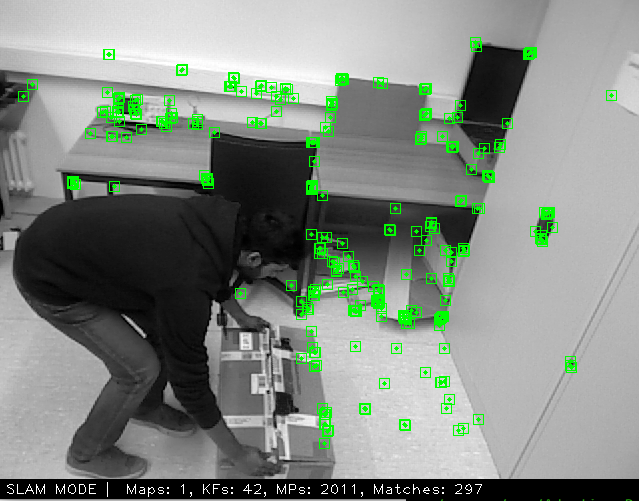}
    \label{fig:KF42} 
  }
  \caption{Scenes in the \texttt{moving\_nonobs\_box} sequence showing a person moving a box between keyframes 29 and 42.} 
  \label{fig:KF_nonobs}
\end{figure}

\begin{figure}[h!]
\centering    \includegraphics[width=.99\linewidth]{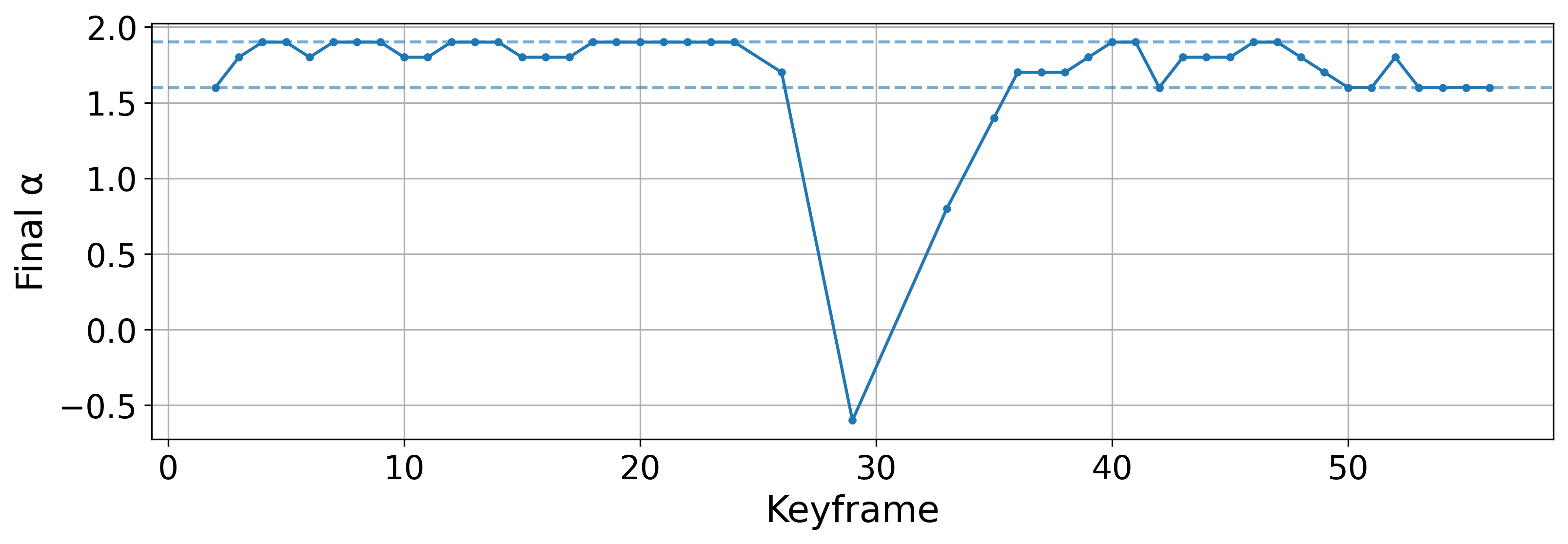}
  \caption{Final $\alpha$ obtained per keyframe in the sequence \texttt{moving\_nonobs\_box} from the Bonn RGB-D Dynamic dataset.} 
  \label{fig:alpha_nonobs}
\end{figure}

Fig.~\ref{fig:KF_nonobs} shows two scenes in the \texttt{placing\_obs\_box} sequence between keyframes 29 and 42. Although the person is detected and filtered, the box is an unknown moving object and has features in it being tracked, which would cause a drift in the estimation. Consistent with this, the adaptive kernel selects a very small $\alpha$ at the keyframe 29, which strongly downweights those outliers, as shown in Fig.~\ref{fig:alpha_nonobs}, where the $alpha$ value values per keyframe as displayed. As the box is moved away, the residual distribution becomes less heavy-tailed, and $\alpha$ rises into the $1.4$--$1.7$ range. Once the object becomes static after keyframe 42, $\alpha$ recovers a value closer to 2 and the estimation returns to near-Gaussian behavior. This example illustrates how the adaptive loss handles previously unseen moving objects without class-specific thresholds.

\subsection{Ablation Studies}

To evaluate the contribution of each component in VAR-SLAM, we analyze three configurations in addition to the full system: ORB-SLAM3 (Huber kernel), VAR-SLAM-RK (kernel-only, replacing Huber with Adaptive Barron loss in Tracking and Local Mapping), and VAR-SLAM-S (semantic-only, filtering people but retaining the Huber kernel). Tab.~\ref{tab:bonn_ablation_sequences} shows the ATE RMSE results for each configuration in two sequences from the Bonn RGB-D Dynamic dataset: \texttt{placing\_nonobs\_box} and \texttt{placing\_nonobs\_box3}.

\begin{figure}[h!]
\centering
  \subfigure[]{%
    \includegraphics[width=.47\linewidth]{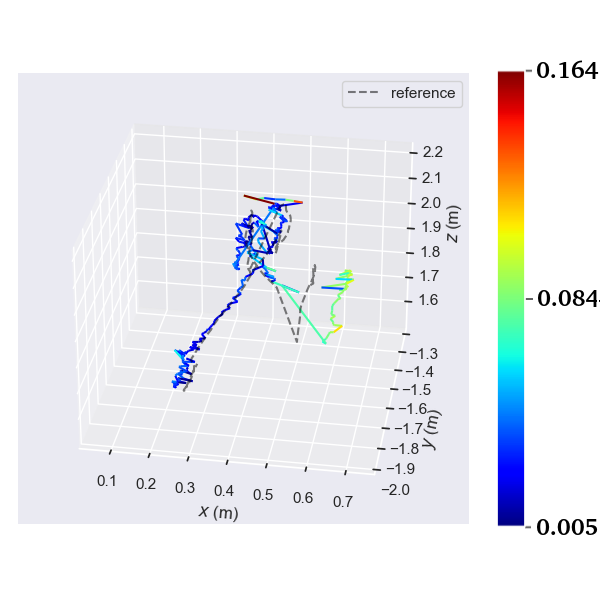}
    \label{nonobsbox3_1}
  } 
  \subfigure[]{%
    \includegraphics[width=.47\linewidth]{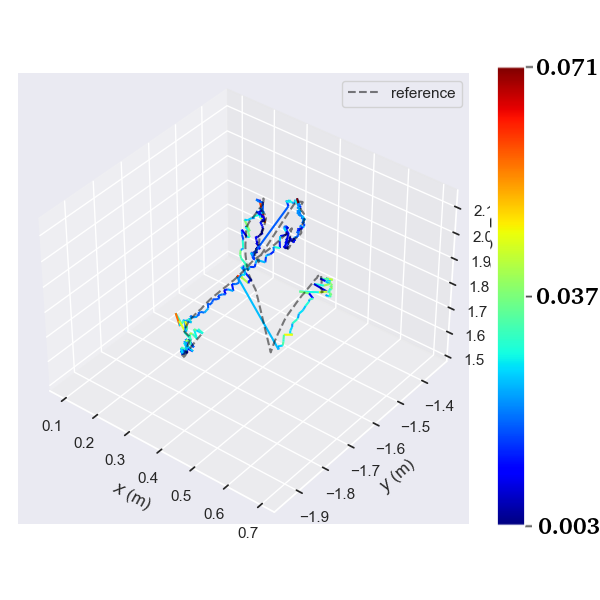}
    \label{nonobsbox3_2} 
  }
  \caption{ Output trajectories in the \texttt{placing\_nonobs\_box3} sequence of (a) VAR-SLAM-SF and (b) VAR-SLAM.} 
  \label{fig:ablation}
\end{figure}

In \texttt{placing\_nonobs\_box3}, the RMSE difference between semantic-only and full VAR-SLAM is modest (4.72 cm against 2.26 cm). However, trajectory plots (Fig. \ref{fig:ablation}) reveal that semantic-only suffers from large local drifts, with maximum ATE exceeding 16 cm. Full VAR-SLAM reduces this worst-case error to $\sim$7 cm. This indicates that while RMSE captures average accuracy, the adaptive kernel also provides significant robustness benefits by suppressing occasional large drifts caused by unknown moving objects.

\begin{table}[h!]
\centering
\caption{Ablation study on the Bonn RGB-D Dynamic dataset. RMSE ATE [cm] for each sequence.}
\label{tab:bonn_ablation_sequences}
\resizebox{\linewidth}{!}{%
\begin{tabular}{lcc}
\hline
 & \multicolumn{2}{c}{Sequence} \\ 
\cline{2-3}
Method & placing\_nonobs\_box & placing\_nonobs\_box3 \\
\hline
ORB-SLAM3       & 80.67 & 16.10 \\
VAR-SLAM-S             & 6.92  & 4.72  \\
VAR-SLAM-RK         & 65.52 & 22.67 \\
Full (VAR-SLAM) & 2.62  & 2.26  \\
\hline
\end{tabular}%
}
\end{table}

\subsection{Run-time analysis}

All our tests were performed in a hardware with a 13th Gen Intel(R) Core(TM) i7-13700H CPU, 16GB RAM, and a
RTX4050 6GB VRAM GPU. With YOLOv4 using GPU, our method runs at 27 FPS on average.

ORB-SLAM3 runs at 13.63 ms per frame on average in our hardware. The average processing time per frame of VAR-SLAM is 14.54 ms, without taking into consideration the detection thread. Thus, the ratio between VAR-SLAM tracking time and ORB-SLAM3 is 1.067. This shows that the implementation of the methodology successfully improved ORB-SLAM3 without compromising its performance.

\subsection{Limitations and Future directions}

We adapt $\alpha$ in Local BA and keep tracking with a fixed $\alpha$ for stability and speed. However, we kept Full BA unchanged. Extending adaptivity to global optimization is a promising direction. Furthermore, our system is susceptible to the problems inherent of vision-only methods, such as textureless environments, and long camera occlusions. A visual-inertial method, for instance, would have superior robustness in these scenarios. Finally, we used the people detection as a proof of concept, but using a light tracker of other known classes could improve the results with minimal overhead.

\section{CONCLUSIONS}

We presented VAR-SLAM, a feature-based visual SLAM system for dynamic environments that combines lightweight semantic filtering of known dynamic classes with a Barron adaptive robust kernel to downweight the influence of unknown moving objects. The shape parameter $\alpha$ is estimated online from the residual distribution in Local BA, enabling the optimizer to transition smoothly between near-Gaussian and heavy-tailed behavior, while the tracking uses a fixed shape value to preserve performance. On TUM RGB-D ``walking'' sequences our method remains competitive with the state-of-the-art. On the Bonn RGB-D Dynamic dataset, especially in scenes with unknown, obstructing moving objects, VAR-SLAM improves both RMSE and worst-case error over strong baselines, all at $\sim$27 FPS. We also report results on two OpenLORIS sequences (market and office), showing transfer without per-dataset tuning. Furthermore, we analyze the evolution of $\alpha$ and show that it naturally converges toward L2 in static segments and lowers in scenes with more outliers, validating the data-driven robustification. We believe this combination of semantic filtering and adaptive robustness offers a practical path to dynamic-scene SLAM without threshold-heavy pipelines. The implementation will be released as open source to support reproducibility and future research.







\bibliographystyle{IEEEtran}
\bibliography{main}


\end{document}